\documentclass{article} 
\usepackage{iclr2016_workshop,times}
\usepackage{url}
\usepackage{amsmath}
\usepackage{amssymb}
\usepackage{amsthm}
\usepackage{graphicx}
\usepackage{pgfplots}
\usepackage{tikz}
\usepackage{tikzscale}
\usepackage{hyperref}

\newcommand{\setN}{\mathbb{N}}
\newcommand{\setZ}{\mathbb{Z}}
\newcommand{\setR}{\mathbb{R}}
\newcommand{\setC}{\mathbb{C}}

\renewcommand{\vec}[1]{\boldsymbol{#1}}

\newcommand{\expn}{\exp^{(n)}}

\usetikzlibrary{external} 

\title{A Differentiable Transition Between \\Additive and Multiplicative Neurons}

\author{Wiebke K\"opp, Patrick van der Smagt\thanks{Patrick van der Smagt is also affiliated with: fortiss, TUM Associate Institute.} \hspace{0.5ex} \& Sebastian Urban\\
Chair for Robotics and Embedded Systems \\
Department of Informatics \\
Technische Universit\"at M\"unchen, Germany \\
\texttt{koepp@in.tum.de, surban@tum.de} 
}

\begin{document}

\maketitle

\begin{abstract}
Existing approaches to combine both additive and multiplicative neural units either use a fixed assignment of operations or require discrete optimization to determine what function a neuron should perform.
However, this leads to an extensive increase in the computational complexity of the training procedure.

We present a novel, parameterizable transfer function based on the mathematical concept of non-integer functional iteration that allows the operation each neuron performs to be smoothly and, most importantly, differentiablely adjusted between addition and multiplication. 
This allows the decision between addition and multiplication to be integrated into the standard backpropagation training procedure.
\end{abstract}

\section{Introduction}

\cite{durbin1989} proposed a neural unit in which the weighted summation is replaced by a product, where each input is raised to a power determined by its corresponding weight.
The value of such a \emph{product unit} is given by
\smash{
$
y_i = \sigma( \prod_j x_j^{W_{ij}} ) \,. 
$}
Using laws of the exponential function this can be written as
\begin{equation}
\vec{y} = \sigma(\exp( W \log \vec{x} ) )  
\label{eq:nnmultmat}
\end{equation}
where $\exp$, $\log$ and $\sigma$ are taken element-wise.
Product units can be combined with ordinary additive units in a hybrid summation-multiplication network.
Yet this poses the problem of how to distribute additive and multiplicative units over the network.
One possibility is to optimize \emph{populations} of neural networks with different addition/multiplication configurations using discrete optimization methods such as genetic algorithms \citep{goldberg1988genetic}.
Unfortunately, these methods require a multiple of the training time of standard backpropagation, since evaluation of the fitness of a particular configuration requires full training of the network.

Here we propose a novel approach, where the distinction between additive and multiplicative neurons is not discrete but \emph{continuous and differentiable}.
Hence the optimal distribution of additive and multiplicative units can be determined during standard, gradient-based optimization.
\section{Continuous interpolation between addition and multiplication}

\paragraph{Functional iteration.}
Let $f: \setR \to \setR$ be an invertible function.
For $n \in \setZ$ we write $f^{(n)}$ for the n-times iterated application of $f$.
Further let $f^{(-n)} = (f^{-1})^{(n)}$ where $f^{-1}$ denotes the inverse of $f$.
We set $f^{(0)}(z) = z$ to be the identity function.
Obviously this definition only holds for integer $n$.

\begin{figure}
\centering
$\vcenter{\hbox{\includegraphics[width=0.32\columnwidth]{psi.tikz}}}$  \hfill
$\vcenter{\hbox{\includegraphics[width=0.32\columnwidth]{expa.tikz}}}$ \hfill
$\vcenter{\hbox{\includegraphics[width=0.32\columnwidth,trim=0 10px 0 0]{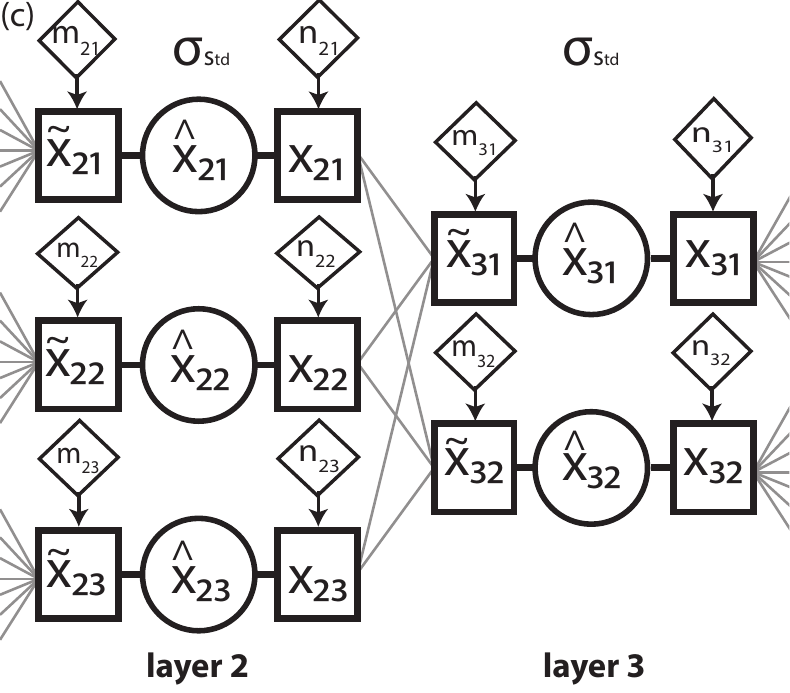}}}$
\caption{
(a) A continuously differentiable solution $\psi(x)$ to Abel's equation \eqref{eq:abel} for the exponential function.
(b) Iterates of the exponential function $\exp^{(n)}(x)$ for $n \in \{-1,-0.9,\dots,0,\dots,0.9,1\}$.
(c) 
A neural network with neurons that can interpolate between addition and multiplication.
In each layer we have
    $\tilde{x}_{li} = \exp^{(m_{li})} ( \sum_j W_{ij} x_{(l-1) j} )$
and $\hat{x}_{li} = \sigma_{\mathrm{std}}(\tilde{x}_{li})$ 
and $ x_{li} = \exp^{(n_{li})} (\hat{x}_{li})$.
Addition occurs for $n_{li} = 0 = m_{(l+1)j}$ and multiplication occurs for $n_{li} = -1$, $m_{(l+1)j} = 1$.
}
\label{fig:abelsol}
\label{fig:abelexp}
\label{fig:adpnet}
\end{figure}

\paragraph{Abel's functional equation.}
Consider the following functional equation given by \cite{abel1826},
\begin{equation}
\psi(f(x)) = \psi(x) + \beta
\label{eq:abel}
\end{equation}
with constant $\beta \in \setC$.
We are concerned with $f(x)=\exp(x)$.
A continuously differentiable solution for $\beta=1$ and $x \in \setR$ is given by
\begin{equation}
\psi(x) = \log^{(k)}(x) + k
\label{eq:abelsol}
\end{equation}
with $k \in \setN \text{ s.t. } 0 \leq \log^{(k)}(x) < 1$.
Note that for $x<0$ we have $k=-1$ and thus $\psi$ is well defined on whole $\setR$.
The function is shown in Fig.~\ref{fig:abelsol}a.
Since $\psi: \setR \to (-1, \infty)$ is strictly increasing, the inverse $\psi^{-1}: (-1, \infty) \to \setR$ exists and is given by
\begin{equation}
\psi^{-1}(\psi) = \exp^{(k)}(\psi - k) \quad 
\label{eq:psiinv}
\end{equation}
with $k \in \setN \text{ s.t. } 0 \leq \psi - k < 1$.
For practical reasons we set $\psi^{-1}(\psi) = -\infty$ for $\psi \leq -1$. 

The derivatives, with the respective definition of $k$ from above, are given by
\begin{equation}
\psi'(x) = \prod_{j=0}^{k-1} \frac{1}{\log^{(j)}(x)}  \,,
\quad \quad \quad
\psi^{-1'}(\psi) = \prod_{j=0}^{k-1} \exp^{(j)}\!\left( \psi^{-1}(\psi - j)  \right)  \,.
\end{equation}

\paragraph{Non-integer iterates of the exponential function.}
By inspection of Abel's equation \eqref{eq:abel}, we see that the $n$th iterate of the exponential function can be written as
\begin{equation}
\exp^{(n)}(x) = \psi^{-1}\!\left( \psi(x) + n \right) \, .
\label{eq:abelexp}
\end{equation}
We are now free to choose $n \in \setR$ and thus \eqref{eq:abelexp} can be seen as a generalization of functional iteration to non-integer iterates.
Hence we can understand the function $\varphi(x) = \exp^{(1/N)}(x)$ as the function that gives the exponential function when iterated $N$ times, see Fig.~\ref{fig:abelexp}b.
Since $n$ is a continuous parameter we can take the derivative of $\exp$ with respect to its argument as well as $n$,
\begin{align}
\exp^{(n')}(x) = \frac{\partial \exp^{(n)}(x)}{\partial n} = \psi^{-1'}\!\left(\psi(x) + n \right) \,,
\quad \, \, \, 
\exp'^{(n)}(x) = \frac{\partial \exp^{(n)}(x)}{\partial x} = \exp^{(n')}(x) \psi'(x) \,.
\nonumber
\end{align}

\paragraph{``Real-valued Addiplication''.}
We define the operator $\oplus_n$ for $x,y \in \setR$ and $n \in \setR$ as
\begin{equation}
x \oplus_n y = \exp^{(n)}\!\left( \exp^{(-n)}(x) + \exp^{(-n)}(y) \right) \, .
\label{eq:addiplication}
\end{equation}
Note that we have $x \oplus_0 y = x + y$ and $x \oplus_1 y = x y$.
For $0 < n < 1$ the operator \eqref{eq:addiplication} interpolates between the elementary operations of addition and multiplication in a continuous and differentiable way.

\paragraph{Neurons that can interpolate between addition and multiplication} can be implemented using a standard neural network by a neuron-dependent, parameterized transfer function,
\begin{equation}
\sigma_{m_{li}}^{n_{li}}(t) = \exp^{(n_{li})}\!\left[ \sigma_{\mathrm{std}}\!\left( \exp^{(m_{li})}(t) \right)  \right] \,,
\label{eq:transfer}
\end{equation}
where $m_{li}, n_{li} \in \setR$ denote parameters specific to a neuron $i$ of layer $l$ and $\sigma_{\mathrm{std}}$ is the standard sigmoid (or any other) nonlinearity.
This corresponds to the architecture shown in Fig.~\ref{fig:adpnet}c.
Since the $m_{li}$s and $n_{li}$s of different layers are not tied together, the network is free to implement arbitrary combinations of iterates of the exponential function.
The operator \eqref{eq:addiplication} occurs as a special case for a pair of neurons with $m_{(l+1)i} = -n_{lj}$.

\section{Experiments}
We examine a synthetic dataset that exhibits multiplicative interactions between inputs.
The function to be approximated is a multi-variate polynomial or multinomial, e.g. 
\begin{equation}
f(x,y) = a_{00} + a_{10}\,x_{1} + a_{01}\,x_{2} + a_{11}\,x_{1}x_{2} + a_{20}\,x_{1}^{2} + a_{02}\,x_{2}^{2}
\label{eq:poly}
\end{equation}
for a multinomial of second degree and two variables $x_{1}$ and $x_{2}$.
Each multinomial can be computed exactly by a three-layer ANN, where the first two layers calculate the products between inputs by using $\log$ and $\exp$ as transfer functions and the final layer uses no transfer function and coefficients $a_{ij}$ as weights. 

\begin{figure}
\centering
\vspace{-14mm}
\includegraphics[width=0.98\columnwidth]{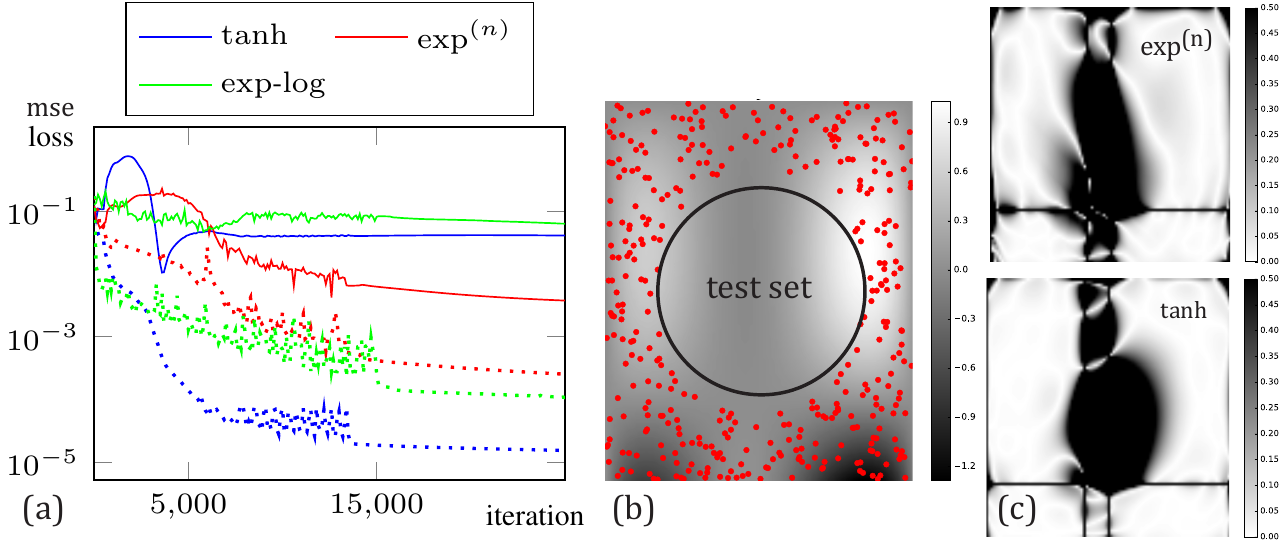}
\tikzexternaldisable
	\caption{
	(a) Test (\protect\tikz[baseline=-\the\dimexpr\fontdimen22\textfont2\relax] \protect\draw (0,0) -- (1em,0);) and training loss (\protect\tikz[baseline=-\the\dimexpr\fontdimen22\textfont2\relax] \protect\draw[dotted,thick] (0,0) -- (1em,0);) loss for approximation of a multivariate polynomial for different net structures. 
	(b) Polynomial data (grayscale coded) to be learned ranging with training points in red. 
	(c) Relative error for networks with $\expn$ and $\tanh$ as transfer functions.}
	\label{fig:mult}
\tikzexternalenable
\end{figure}

We randomly sample 600 data points from a multinomial of two variables $x_{1}$ and $x_{2}$ and degree four with $x_{1}, x_{2} \in [0,1]$.
In order to analyze long-range generalization performance, the dataset is split into test and training set in a non-random way. All points within a circle of radius $r=0.33$ around $(0.5, 0.5)$ constitute the test set, while all other points make up the training set, see Fig.~\ref{fig:mult}b. 

We train three different three-layer net structures optimized by Adam \citep{kingma2015}, where the final layer is additive with no additional transfer function in all cases. 
The first neural network is purely additive with $\tanh$ as the transfer function for the first two layers. 
The second network uses \eqref{eq:transfer} in the first two layers, thus allowing interpolation between addition and multiplication. 
In this configuration we set $\sigma_{\mathrm{std}}(x) = x$.
The third and final network uses $\log$ and $\exp$ as transfer functions in the first two layers, yielding fixed multiplicative interaction between the inputs.
All weights of all networks, including the transfer function parameters of our model, are initialized with zero mean and variance $10^{-2}$.
Hence, our model starts with a mostly additive configuration of neurons.
The progression of training and test loss for all three structures is displayed in Fig.~\ref{fig:mult}a. 

As the relative error of the approximation (see Fig.~\ref{fig:mult}c) shows, our proposed transfer function generalizes best in this experiment.  
Surprisingly, our model even surpasses the $\log$-$\exp$ network, which perfectly resembles the data structure due to its fixed multiplicative interactions in the first two layers.
We hypothesize that training a neural network with multiplicative interactions but otherwise randomly initialized weights is hindered by very complex error landscapes that make it difficult for gradient-based optimizers to escape deep local minima.
Our model seems unaffected by this issue.
We suspect that, since it starts with additive interactions, it can find reasonable values for the weights before moving into the multiplicative regime.

\section{Conclusion}
We proposed a method to differentiably interpolate between addition and multiplication and showed how it can be integrated into standard neural networks by using a parameterizable transfer function.
Here we limited ourselves to the real domain, thus multiplication can only occur between two positive values since $\log x = -\infty$ for $x\leq0$.
An extension of this framework to the complex domain proposed by \cite{2015arXiv150305724U} eliminates this restriction but doubles the number of weights.

\section*{Acknowledgments}
This project was funded in part by the German Research Foundation (DFG) SPP 1527 Autonomes Lernen and by the TACMAN project, EC Grant agreement no.\ 610967, within the FP7 framework programme.
\nocite{KuczmaChoczewskiGer199007}

\bibliography{addiplication-iclr}

\begin{thebibliography}{6}
\providecommand{\natexlab}[1]{#1}
\providecommand{\url}[1]{\texttt{#1}}
\expandafter\ifx\csname urlstyle\endcsname\relax
  \providecommand{\doi}[1]{doi: #1}\else
  \providecommand{\doi}{doi: \begingroup \urlstyle{rm}\Url}\fi

\bibitem[Abel(1826)]{abel1826}
N.H. Abel.
\newblock Untersuchung der functionen zweier unabh\"{a}ngig ver\"{a}nderlichen
  gr\"{o}{\ss}en x und y, wie f(x, y), welche die eigenschaft haben, da{\ss}
  f(z, f (x, y)) eine symmetrische function von z, x und y ist.
\newblock \emph{Journal f\"{u}r die reine und angewandte Mathematik},
  1826\penalty0 (1):\penalty0 11--15, 1826.

\bibitem[Durbin \& Rumelhart(1989)Durbin and Rumelhart]{durbin1989}
Richard Durbin and David~E Rumelhart.
\newblock {Product Units: A Computationally Powerful and Biologically Plausible
  Extension to Backpropagation Networks}.
\newblock \emph{Neural Computation}, 1:\penalty0 133--142, 1989.

\bibitem[Goldberg \& Holland(1988)Goldberg and Holland]{goldberg1988genetic}
David~E Goldberg and John~H Holland.
\newblock Genetic algorithms and machine learning.
\newblock \emph{Machine learning}, 3\penalty0 (2):\penalty0 95--99, 1988.

\bibitem[Kingma \& Ba(2015)Kingma and Ba]{kingma2015}
Diederik Kingma and Jimmy Ba.
\newblock Adam: A method for stochastic optimization.
\newblock \emph{ArXiv e-prints}, July 2015.
\newblock URL \url{http://arxiv.org/abs/1412.6980}.

\bibitem[Kuczma et~al.(1990)Kuczma, Choczewski, and
  Ger]{KuczmaChoczewskiGer199007}
Marek Kuczma, Bogdan Choczewski, and Roman Ger.
\newblock \emph{Iterative Functional Equations (Encyclopedia of Mathematics and
  its Applications)}.
\newblock Cambridge University Press, 1st edition, July 1990.

\bibitem[{Urban} \& {van der Smagt}(2015){Urban} and {van der
  Smagt}]{2015arXiv150305724U}
S.~{Urban} and P.~{van der Smagt}.
\newblock {A Neural Transfer Function for a Smooth and Differentiable
  Transition Between Additive and Multiplicative Interactions}.
\newblock \emph{ArXiv e-prints}, March 2015.
\newblock URL \url{http://arxiv.org/abs/1503.05724}.

\end{thebibliography}
\bibliographystyle{iclr2016_workshop}

\end{document}